\documentclass[11pt,a4paper]{article}
\usepackage[hyperref]{emnlp2020}
\usepackage{url}
\usepackage{times}
\usepackage{xcolor}
\usepackage{amsmath}
\usepackage{arydshln}
\usepackage{latexsym}
\usepackage{booktabs}
\usepackage{enumitem}
\usepackage{multirow}
\usepackage{graphicx}
\usepackage{subcaption}

\usepackage{microtype}

\aclfinalcopy 

\title{Knowledge Grounded Conversational Symptom Detection\\ with Graph Memory Networks}

\author{Hongyin Luo$^1$ $    $ Shang-Wen Li$^2$\thanks{Work done before the second author joined Amazon} $    $ James Glass$^1$ \\
  $^{1}$MIT CSAIL \\
  $^{2}$Amazon AI\\ 
  {\tt hyluo@mit.edu, shangwel@amazon.com, glass@mit.edu}\\
  }

\date{}

\begin{document}
\maketitle
\begin{abstract}
In this work, we propose a novel goal-oriented dialog task, automatic symptom detection. We build a system that can interact with patients through dialog to detect and collect clinical symptoms automatically, which can save a doctor's time interviewing the patient. Given a set of explicit symptoms provided by the patient to initiate a dialog for diagnosing, the system is trained to collect implicit symptoms by asking questions, in order to collect more information for making an accurate diagnosis. After getting the reply from the patient for each question, the system also decides whether current information is enough for a human doctor to make a diagnosis. To achieve this goal, we propose two neural models and a training pipeline for the multi-step reasoning task. We also build a knowledge graph as additional inputs to further improve model performance. Experiments show that our model significantly outperforms the baseline by 4\%, discovering 67\% of implicit symptoms on average with a limited number of questions.
\end{abstract}

\section{Introduction}
In a typical clinical conversation between a patient and a doctor, the patient initiates the dialog by providing a number of explicit symptoms as a self-report. Based on this information, the doctor asks about other possible symptoms, in order to make an accurate diagnosis and suggest treatments. This is a multi-step reasoning process. At each step, the doctor choose a symptom to ask or concludes the diagnosis by considering the dialog history and possible diseases.

With recent advances in deep reinforcement learning \citep{mnih2013playing} and task-oriented dialog systems \citep{bordes2016learning,wen2016network}, recent studies have proposed human-computer dialog systems for automatic diagnosis \citep{wei2018task}. The automatic diagnosis system applied a deep Q network (DQN) to decide whether to continue the dialog by asking about a symptom or conclude the diagnosis by predicting a disease. \citet{xu2019end} proposed a knowledge-routed DQN that improves this process by considering relations among diseases and symptoms. The systems described above can achieve around 70\% accuracy in making a diagnosis among 4 common diseases and detects a few implicit symptoms.

However, the automatic diagnosis systems is far from being ready for clinical diagnosis, since there is still a gap in accuracy between the system and human doctors. Furthermore, he current legislation system has to be amended such that liability can be clarified when the system mis-diagnoses. As a result, current machines are not ready to replace human doctors, but they can still detect symptoms automatically to assist doctors making decision more efficiently.

In this work, we propose a system that automatically detects clinical symptoms of a patient through dialog, instead of diagnosing. With such symptom information collected, a doctor can be more knowledgeable before interviewing the patient. Only those symptoms not covered by the dialog system should be asked during the interview. Thus both the time and cost of the clinical interviews can be reduced. Meanwhile, the diagnosis is still made by the doctor, which guarantees the accuracy and responsibility.

Table \ref{tab:example} shows two example conversations generated by a dialog agent for automatic diagnosis (AD) and a dialog agent for automatic symptom detection (ASD). AD and ASD agents conclude the conversations differently. The goal of the AD system is making the correct diagnosis, so it concludes the conversation with a prediction of the possible disease. The ASD system aims at discovering as many implicit symptoms as possible, thus it stops the conversation when after deciding that current information is enough for a doctor to make diagnosis. As shown in Table \ref{tab:example}, the number of turns of the ASD system is possibly more than an AD system, and it covers more implicit symptoms that are not mentioned in the patient's self-report.
\begin{table}[]
\begin{tabular}{@{}ll@{}}
\toprule
\multicolumn{2}{c}{\textbf{I. Automatic Diagnosis (AD)}}            \\ \midrule
\textbf{User:}  & The baby has a \textcolor{blue}{Runny Nose}.       \\
\textbf{Agent:} & Does the baby \textcolor{red}{cough}?                          \\
\textbf{User:}  & Yes, the baby is coughing.                \\
\textbf{Agent:} & Does the baby have a \textcolor{green}{fever}?                     \\
\textbf{User:}  & I am not sure.                            \\
\textbf{Agent:} & It is upper respiratory infection \\ \midrule
\multicolumn{2}{c}{\textbf{II. Automatic Symptom Detection (ASD)}}            \\ \midrule
\textbf{User:}  & The baby has a \textcolor{blue}{Runny Nose}.       \\
\textbf{Agent:} & Does the baby \textcolor{red}{cough}?                          \\
\textbf{User:}  & Yes, the baby is coughing.                \\
\textbf{Agent:} & Does the baby \textcolor{red}{sneeze}?                          \\
\textbf{User:}  & Yes, the baby is sneezing.                \\
\textbf{Agent:} & Does the baby have a \textcolor{green}{fever}?                     \\
\textbf{User:}  & I am not sure.                            \\
\textbf{Agent:} & Does the baby have a \textcolor{red}{headache}?                     \\
\textbf{User:}  & Yes, the baby has a headache.                            \\
\textbf{Agent:} & Does the baby have \textcolor{red}{Phlegm}?                     \\
\textbf{User:}  & Yes, the baby has Phlegm.                            \\
\textbf{Agent:} & \begin{tabular}[c]{@{}l@{}} Thank you for the information!\\ A report has been sent to your doctor.\end{tabular} \\ \bottomrule
\end{tabular}
\caption{Two examples of dialog between different systems and a user. Conversation I is generated by an automatic diagnosis system, and conversation II is generated by an automatic symptom detection system. The explicit symptom is highlighted in blue, the implicit symptoms are highlighted in red, and unrelated symptoms are marked in green.}
\label{tab:example}
\end{table}

In this work, we focus on the conversational ASD task. We propose a system that predicts implicit symptoms and whether to conclude the conversation with neural networks. To train the neural networks, we borrow the idea of the masked language model \citep{devlin2018bert} and simulate both training and test datasets. To improve the performance of the system, we annotate a medical knowledge graph based on an online medical dictionary. Then we propose a graph memory network (GMemNN) architecture to utilize the external knowledge graph. We also propose two metrics: symptom hit rate and unrelate rate to evaluate the performance of the system. 

We make following contributions in this paper,
\begin{itemize}[noitemsep,topsep=0pt]
\item We propose the conversational symptom detection task and evaluation metrics.
\item We annotate a knowledge graph in the medical domain to enrich the current corpus.
\item We propose a graph memory network (GMemNN) architecture to build the dialog agent, which produces the state-of-the-art performance.
\end{itemize}

\section{Related Work}
\subsection{Task-Oriented Dialog Systems}
Task-oriented dialog systems aim at completing a specific task by interacting with users through natural language, and the main challenge is learning a dialog policy manager \citep{papineni2001natural}. Typical applications include flight booking \citep{seneff2000dialogue}, movie recommendation \citep{dodge2015evaluating,fazelzarandi2017learning}, restaurant reservation \citep{bordes2016learning}, and vision grounding \citep{chattopadhyay2017evaluating}. Recently, such systems have been applied in automatic diagnosis \citep{wei2018task,xu2019end,luo2020prototypical}. The authors of \citet{de2017guesswhat} proposed the GuessWhat game, which requires computers to guess a visual object given a natural language description by asking a series of questions. The GuessWhat game is similar with our task in the medical domain.


\subsection{Knowledge and Graph Processing}
Many tasks require processing knowledge in different formats. \citet{sukhbaatar2015end} proposed memory networks (MemNNs) for question answering. The context of the question, or knowledge, is stored in an external memory bank and the model reads information from the memory with an attention mechanism. The MemNN model is also applied in question answering in the movie domain \citep{miller2016key}, video question answering \citep{luo2019integrating}, and stance detection \citep{mohtarami2018automatic}. The neural Turing machine \citep{graves2014neural} and the neural computer \citep{graves2016hybrid} also applied external memory banks, and enable the models to write into and read from the external memory cells dynamically.

In many tasks, knowledge can be organized as graphs. Recent studies have proposed different neural models for processing graph-structured data. The graph neural networks (GNNs) \citep{scarselli2008graph} uses neural networks to perform message propagation on graphs. The graph convolutional networks (GCNs) \citep{kipf2016semi} employed a multi-layer architecture to learn node embeddings by integrating the information of the nodes and their neighbors. The graph attention networks \citep{velivckovic2017graph} integrates node embeddings with an attention mechanism. \citet{shang2019gamenet} proposed a graph augmented memory network (GAMENet) model for medication recommendation. A similar idea that combines graphs and memory networks is proposed in \citet{pham2018graph} for molecular activity
prediction. In this work, we also propose a memory network architecture that processes graph-structured knowledge, but focus on bipartite graphs.

\section{Data and Task Definition}
In this section, we formally define the automatic symptom detection task and describe the corpus used to train and evaluate the model. We first introduce the Muzhi corpus \citep{wei2018task}, then describe the task based on the corpus. Lastly, we describe the medical knowledge graph we annotated and the annotation method.

\subsection{Muzhi Corpus}
We train and evaluate our models using the Muzhi corpus. The corpus was collected from a online medical forum\footnote{\url{http://muzhi.baidu.com}}, including 4 common diseases and 66 symptoms. The corpus contains 710 dialog sessions represented as 710 user goals. Each user goal includes a set of explicit symptoms as the user's self report, and a set of implicit symptoms queried by doctors. An example of a user goal is shown in Table \ref{tab:muzhi}.

In the example, 1 means that the patient confirms a symptom, while 0 means that the patient is confident that the symptom does not exist. Other symptoms not listed in the user goal are considered either unrelated to the diagnosis, or the patient is not sure about their existence. In the Muzhi corpus, each user goal contains $2.35$ explicit symptoms and $3.26$ implicit symptoms on average.
\begin{table}[]
\centering
\begin{tabular}{@{}lll@{}}
\toprule
\textbf{Disease\_tag} & Bronchiolitis       &           \\ \midrule
\textbf{Exp Sym}   & Runny Nose: 1  & Cough: 1  \\
\textbf{Imp Sym}   & Sore Throat: 1            & Emesis: 0 \\
\textbf{}            & Harsh Breath: 1 & Fever: 0 \\           \bottomrule
\end{tabular}
\caption{An example of a user goal in the Muzhi corpus, containing explicit symptoms and implicit symptoms. 1 means a symptom is confirmed by the patient, while 0 means that a symptom is denied by the patient.}
\label{tab:muzhi}
\end{table}

\subsection{Automatic Conversational Symptom detection Task}
The goal of the automatic conversational symptoms detection (ASD) task is detecting as many implicit symptoms as possible through dialogs with the patients, limiting the number of dialog turns. The initial input of a dialog agent is the set of explicit symptoms. Based on the query and user response of each step, the system decides a new symptom to ask, or stop the dialog.

All implicit symptoms, including the positive and negative ones, are considered as the target of the system. The user goals are collected from real doctor-patient conversation, so we consider every queried symptom a necessary step of making an accurate diagnosis. The systems are evaluated with two metrics. We say model A outperforms model B if model A discovers more implicit symptoms, and queries less unrelated symptoms.

\subsection{Medical Knowledge Graph}
We annotate a medical knowledge graph to provide information about the relations among symptoms and diseases based on the symptoms included in the Muzhi corpus. As described above, we have 66 symptoms in total. We regard each symptom and disease as a node in the graph and annotate symptom-symptom and symptom-disease edges based on the A-Hospital\footnote{\url{http://www.a-hospital.com/}} website, which contains webpages for both symptoms and diseases.

We propose a novel annotation method to build the medical knowledge graph considering complications. The symptom pages in A-Hospital describes a series of diseases that can cause a symptom. Meanwhile, it also listed most possible symptoms to appear if the target symptom is caused by a certain disease. We regard these symptoms as complications and make use of this information. In practice, we annotate the knowledge graph with the following method,
\begin{itemize}[noitemsep,topsep=0pt]
\item[1.] For each symptom $s$ and its related disease $d$, add edge $(s, d)$. 
\item[2.] For each symptom $s$, its related disease $d$, and complication $c$, add edge $(s, c)$.
\end{itemize}
\begin{figure}[t]
\centering
\includegraphics[height=2.08in]{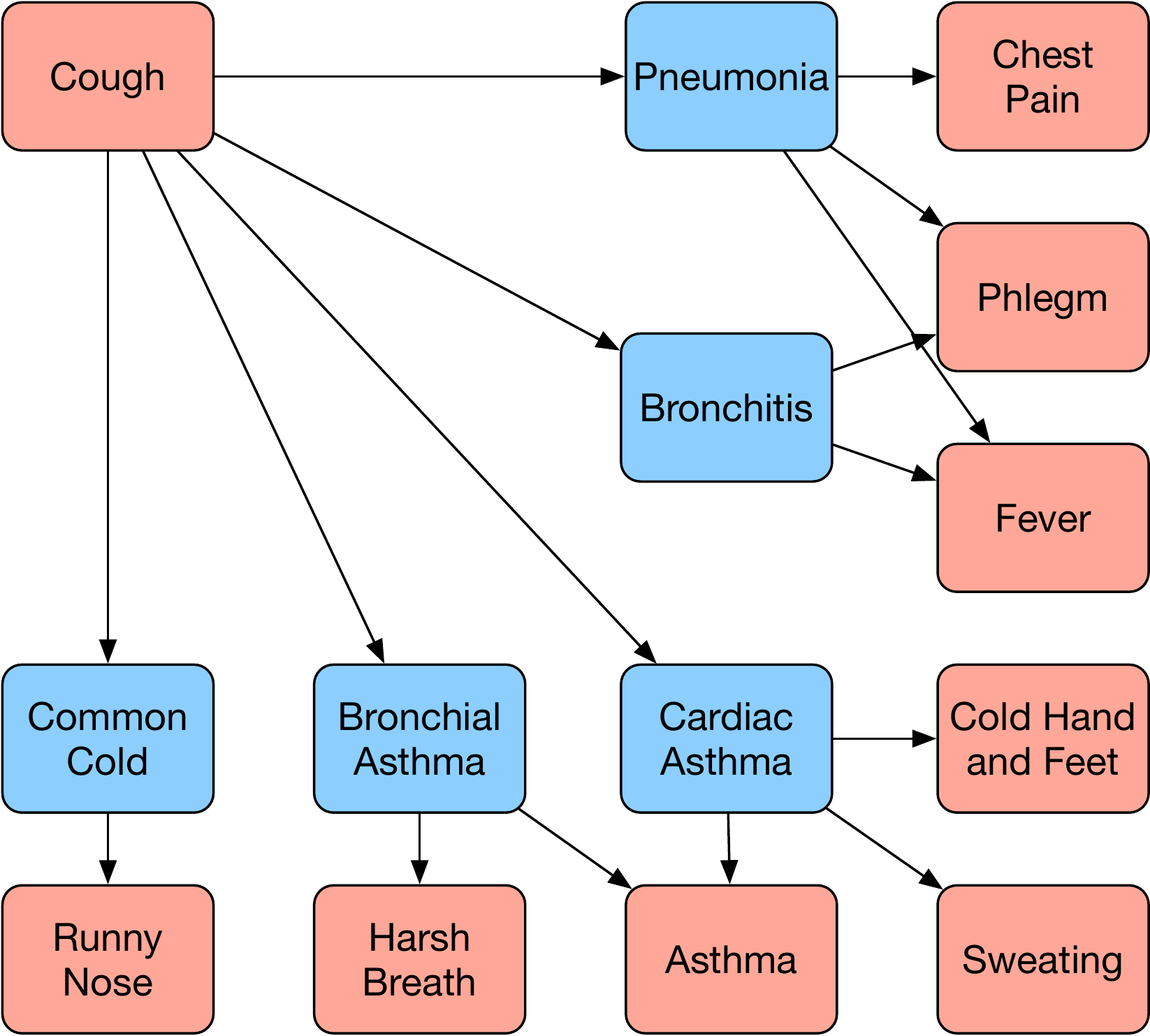}
\caption{An example of an annotated symptom in the knowledge graph. Red blocks represent symptoms and blue blocks stands for disease. ``Cough'' is the target symptom and other symptoms are complications.}
\label{fig:kb}
\end{figure}

An example of the annotated knowledge is shown in Figure \ref{fig:kb}, and Table \ref{tab:kb-stats} summaries the knowledge graph we annotated. In the table, S-D edge stands for symptom-disease edge and S-C edge stands for symptom-disease-symptom edge. The number of S-C edge is lower than multiplying the number of symptoms per disease and diseases per symptom is that only a subset of symptoms caused by a disease are regarded as significant complications of a given symptom.
\begin{table}[]
\centering
\begin{tabular}{@{}lc@{}}
\toprule
\textbf{Items}        & \textbf{Statistics} \\ \midrule
Num. Sym.             & 66                  \\
Num. Dise.            & 28                  \\
Num. Edge             & 1094                \\
Num. S-D Edge         & 284                 \\
Num. S-C Edge       & 810                \\
\bottomrule
\end{tabular}
\caption{A statistics of the annotated knowledge graph of symptoms, diseases, and complications. Note that both symptom-disease and symptom-complication edges exist.}
\label{tab:kb-stats}
\end{table}

\section{Methods}
In this section, we introduce the structure and pipeline of the proposed automatic symptom detection system, including dialog state representation, the neural models for predicting symptoms and dialog actions, the training strategy, and the evaluation metrics.

\subsection{Dialog State Representation}
Automatic symptom detection is a multi-step reasoning task handled by action and symptom predictions. Both tasks are accomplished with neural networks based on the current dialog state.

The first step of building such a system is representing dialog states with vectors that can be processed by the neural networks. Following the method applied in \citet{wei2018task} for vectorizing the dialog states, each dialog state consists of 4 parts:

\noindent \textbf{I. UserAction}: The user action of the previous dialog turn. Possible actions are:
\begin{itemize}[noitemsep,topsep=0pt]
\item \textbf{SelfReport}: A user sends a self-report containing a set of explicit symptoms.
\item \textbf{Confirm}: A user confirms that a queried symptom exists.
\item \textbf{Deny}: A user indicates that a queried symptom does not exist.
\item \textbf{NotSure}: A user replies ``not sure'' when an unrelated symptom is queried.
\end{itemize}
\textbf{II. AgentAction}: The previous action of the dialog agent. Possible actions are:
\begin{itemize}[noitemsep,topsep=0pt]
\item \textbf{Initiate}: The system initiate the dialog and ask the user to send the self-rport.
\item \textbf{Request}: The system query about the existence of a symtom.
\end{itemize}
\textbf{III. Slots}: Contains all symptoms appeared in the dialog history and their status. Each symptom has 4 possible status,
\begin{itemize}[noitemsep,topsep=0pt]
\item \textbf{Confirmed}: Confirmed by the user.
\item \textbf{Denied}: Denied by the user.
\item \textbf{Unrelated}: The symptom is not necessary for the doctor to make an accurate diagnosis.
\item \textbf{NotQueried}: A symptom has not been queried by the agent.
\end{itemize}
\textbf{IV. NumTurns}: Indicates the length of the dialog history, in other words, current number of turns.

In each step, only one value is selected for UserAction, AgentAction, and NumTurns, and we represent them with one-hot vectors $a^u, a^r$, and $n$ respectively. We use a 66-dimension vector $s$ to represent the Slots, where each dimension indicates the status of a symptom. If a symptom is confirmed, the corresponding dimension is set to $1$. If a symptom is denied, the corresponding dimension is set to $-1$. If a symptom is unrelated to the diagnosis process, and the dimension is set to $-2$. All other dimensions are set to $0$. The final input of the neural networks at the $t$-th step is represented as
\begin{equation}
x_t = [a_t^u, a_t^r, n_t, s_t]
\end{equation}
which is genereted by concatenating all the vectors decribed above
\begin{figure*}[t!]
    \centering
    \begin{subfigure}[t]{0.23\textwidth}
        \centering
        \includegraphics[height=1in]{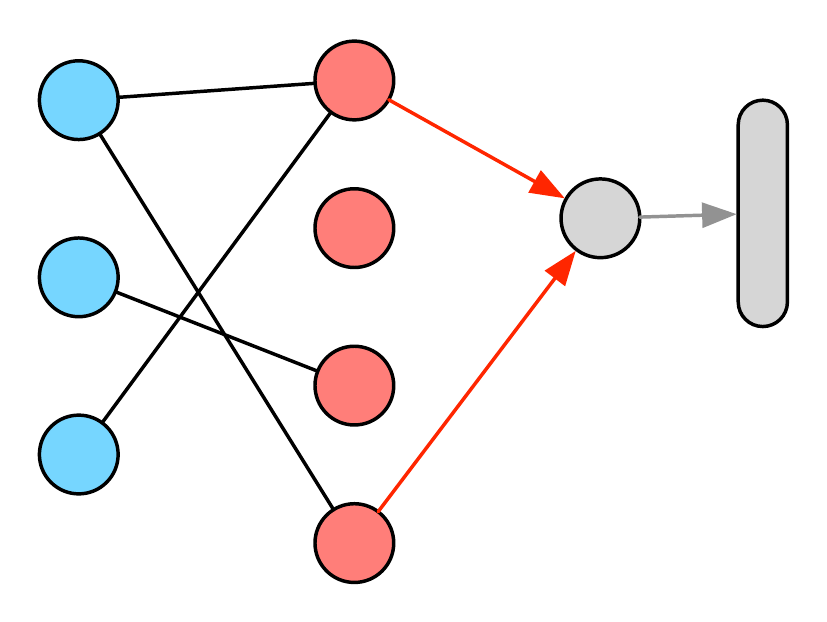}
        \caption{Initiate patient embedding with edges with input slots.}
        \label{fig:step:1}
    \end{subfigure}
    ~ 
    \begin{subfigure}[t]{0.23\textwidth}
        \centering
        \includegraphics[height=1in]{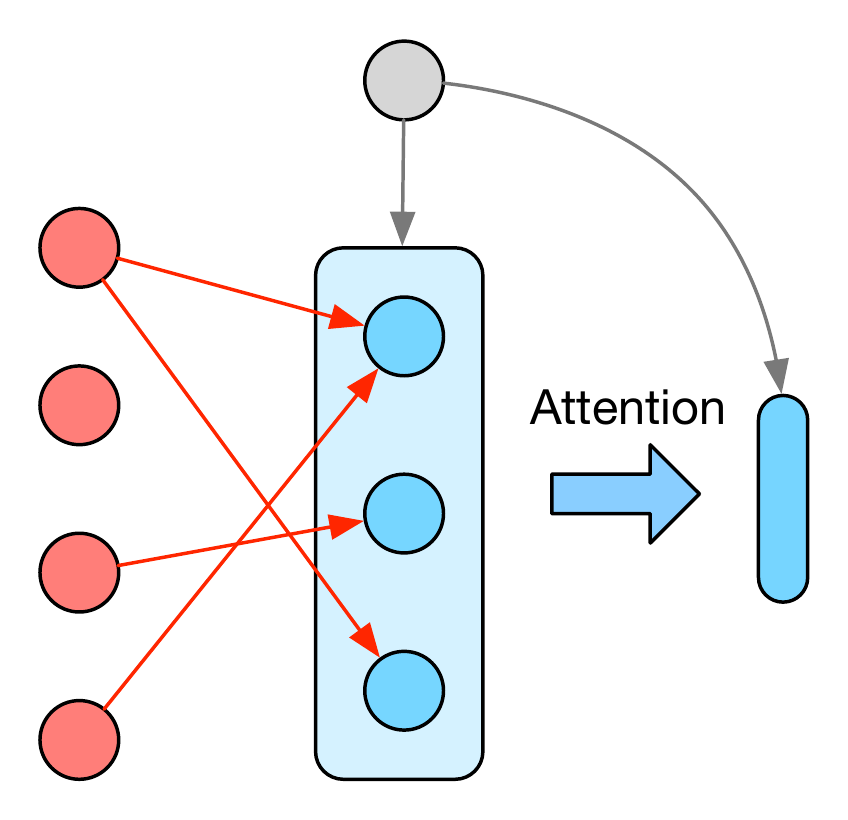}
        \caption{Integrate disease information with attention.}
        \label{fig:step:2}
    \end{subfigure}
    ~ 
    \begin{subfigure}[t]{0.23\textwidth}
        \centering
        \includegraphics[height=1in]{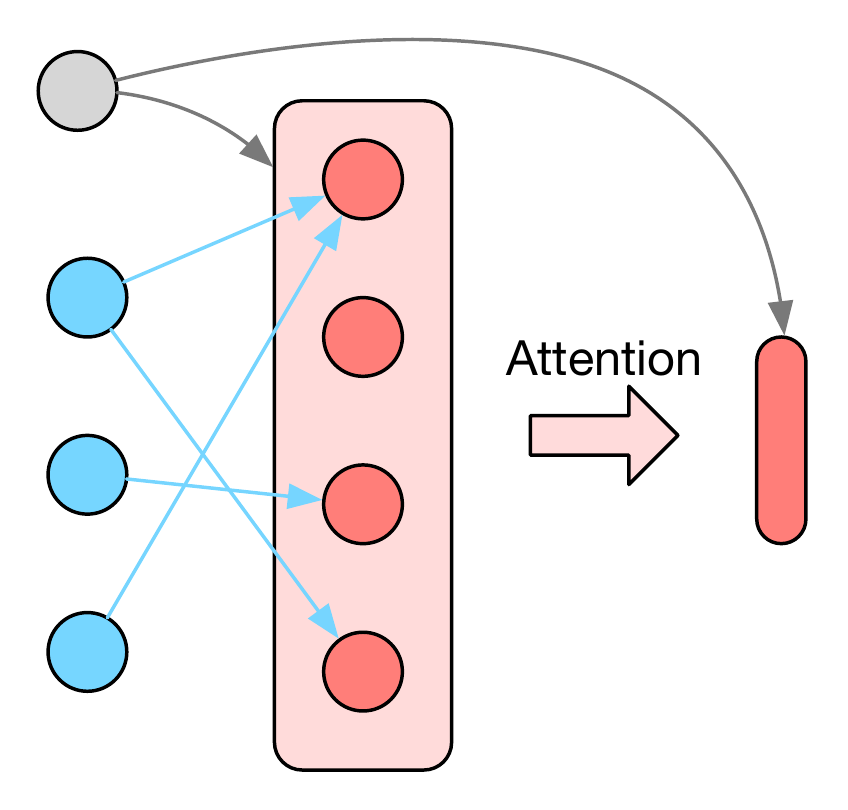}
        \caption{Integrate symptom information with attention.}
        \label{fig:step:3}
    \end{subfigure}
    ~
    \begin{subfigure}[t]{0.25\textwidth}
        \centering
        \includegraphics[height=1in]{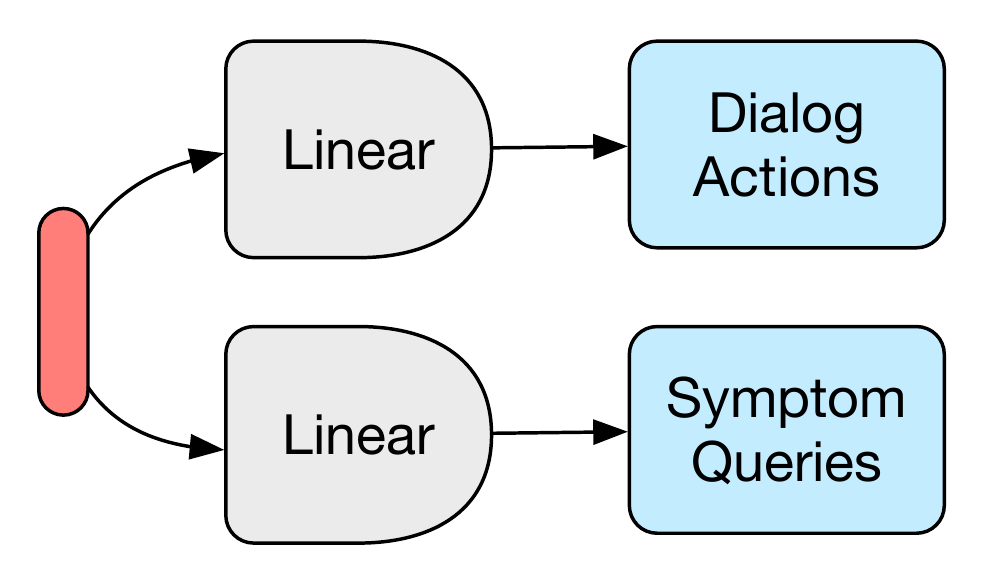}
        \caption{Predict action and symptom with linear transformations.}
        \label{fig:step:4}
    \end{subfigure}
    
    \caption{The 4 steps for processing an input dialog state with a graph memory network (GMemNN). The gray nodes stand for patient, the red nodes represent symptoms, and the blue nodes represent diseases. The edges with arrows, which are labeled with same color as their source nodes, indicate the direction of message propagation.}
    \label{fig:gmemnn}
\end{figure*}

\subsection{Models}
\subsubsection{Multi-Layer Perceptrons}
The first neural model we apply in this work is a multi-layer perceptron (MLP) with 1 hidden layer. The same neural network is applied in \citet{wei2018task} and \citet{xu2019end} for the automatic diagnosis task. With input $x$, the feed forward process of the MLP is shown as follows,
\begin{equation}
\begin{split}
& h = ReLU(W_1 \cdot x + b_1) \\
& y = Softmax(W_2 \cdot h + b_2)
\end{split}
\end{equation}
where Softmax calculates probabilistic distribution by
\begin{equation}
Softmax(a_i) = \frac{e^{a_i}}{\sum_j e^{a_j}}
\end{equation}
The MLP is used for both implicit symptom and dialog action predictions. Note that the MLP model only uses the dialogs in the training set, and does not use the knowledge graph we annotated.

\subsubsection{Graph Memory Networks}
Limited by the structure, MLPs cannot directly utilize the knowledge graph, which contains necessary medical knowledge for clinical diagnosis. Inspired by previous studies on processing knowledge and graphs \citep{sukhbaatar2015end,velivckovic2017graph}, we propose graph memory networks (GMemNN) that utilizes the medical knowledge graph to improve the performance of the automatic symptom detection system.

The knowledge graph is stored in an external memory bank. In each step, we regard a patient as a node connected with the known symptoms in the graph. Our purpose is to learn the embedding of the patient node and predict dialog actions and symptoms based on it. The prediction using GMemNN contains 4 steps: 1. encoding dialog states, 2. integrating potential disease information, 3. integrating complication symptoms, and 4. predicting action/symptom. The 4 steps are illuminated in Figure \ref{fig:gmemnn}.

\noindent \textbf{Dialog State Encoding} The GMemNN encodes the input dialog states with a lookup matrix, or a linear transformation. Given an input dialog state representation $x$, the network encodes the dialog state with
\begin{equation}
u^0 = W_x \cdot x + b_x
\end{equation}
Note that no non-linear activation is applied on $u$ at this step, and $u$ is considered as the initial embedding of the patient node in the graph.

\noindent \textbf{Integrating Disease Information} After encoding the dialog state, we update the patient embedding using the embeddings of possible diseases. We calculate an embedding to summarize potential diseases using the attention mechanism for reading from the memory bank applied in the memory networks \citep{sukhbaatar2015end}.

Similar with the method applied in the MemNN, we first calculate two sets of embeddings for the diseases based on their neighbors, or related symptoms, in the knowledge graph. In this paper, we use $W_m^s$ to denote the symptom embedding matrices for calculating attentions on memory, and $W_c^s$ to denote the symptom embeddings for calculating outputs. The related symptoms are summarized with the adjacency matrix $A_d$ between symptoms and diseases.
\begin{equation}
\begin{split}
& d_{i, m}^1 = d_{i, m}^0 + A_d^i \* W_m^s \* D_{d, i}^{-1} \\
& d_{i, c}^1 = d_{i, c}^0 + A_d^i \* W_c^s \* D_{d, i}^{-1}
\end{split}
\end{equation}

where $d_{i, \cdot}^1$ represents the updated embedding of disease $i$, $d_{i, \cdot}^0$ is the initial disease embedding, $W_\cdot^s$ stands for the symptom embedding matrix for updating disease embeddings. $A_d^i$ is the $i$-th row of $A_d$, and $D_{d,i}$ is the disease node degree for normalization. This is a 
variant of the normalization method proposed in \citet{kipf2016semi}.

Then we summarize potential diseases using $d_m^1$, $d_c^1$, and the initial input embedding $u^0$.
\begin{equation}
\begin{split}
& e^d = \sum_i \alpha_i^d \cdot d_{i, c}^1 \\
& \alpha_i^d = Softmax(u^0 \cdot d_{i, m}^1)
\end{split}
\end{equation}

Then we update the initial patient embedding $u^0$ by integrating disease embeddings.
\begin{equation}
u^d = ReLU(u^0 + e^d)
\end{equation}

\noindent \textbf{Integrating Symptom Information} After integrating the information of possible diseases, the model continues integrating the complication symptom information to produce the final patient embedding. For symptom $i$, given the initial symptom embeddings $s_{i, \cdot}^0$, the adjacency matrix $A_s$ between symptom and symptom, we calculate symptom embeddings with
\begin{equation}
\begin{split}
& s_{i, m}^1 = s_{i, m}^0 + A_s^i \* W_m^s \* D_{s, i}^{-1} + A_d^{\cdot, i} \* W_m^d \* D_{d, \cdot, i}^{-1} \\
& s_{i, c}^1 = s_{i, c}^0 + A_s^i \* W_c^s \* D_{s, i}^{-1} + A_d^{\cdot, i} \* W_c^d \* D_{d, \cdot, i}^{-1}
\end{split}
\end{equation}

where $W_\cdot^s$ is the complication symptom embedding matrix, $W_\cdot^d$ is the disease embedding matrix. $D_{s, i}$ is the number of neighbor symptoms of symptom $i$, and $D_{d, i}$ is the number of neighbor diseases of symptom $i$.

Similarly, we summarize the complication symptoms by
\begin{equation}
\begin{split}
& e^s = \sum_i \alpha_i^s \cdot s_{i, c}^1 \\
& \alpha_i^s = Softmax(u^d \cdot s_{i, m}^1)
\end{split}
\end{equation}

Then we get the final patient embedding by integrating $u^d$ with the complication symptoms embedding
\begin{equation}
u^{d, s} = ReLU(u^d + e^s)
\end{equation}

$u^{d, s}$ stands for a patient embedding that has integrated both disease and symptom information.

\noindent \textbf{Action/Sympotom Prediction} The GMemNN model predicts both dialog actions and symptoms with linear transformations based on the same patient embedding $u^{d, s}$.
\begin{equation}
\begin{split}
& y^{act} = W^{act} \cdot u^{d, s} + b_{act} \\
& y^{sym} = W^{sym} \cdot u^{d, s} + b_{sym}
\end{split}
\end{equation}

The action and symptom distributions are calculated with $y^{act}$ and $y^{sym}$ with the Softmax function. The available dialog actions are Conclude and Query, and the prediction space of the symptom prediction network is the 66 symptoms except the known symptoms.

\subsubsection{Training}
The Muzhi dataset does not contain any dialog history to mimic. Inspired by the masked language model training pipeline proposed by \citet{devlin2018bert}, we construct our own training set by randomly masking and sampling symptoms.

\noindent \textbf{Symptom Prediction} We build the training set by simulating dialog states from user goals in the original training set of the Muzhi corpus. We consider user goal $g_i$ with explicit symptom set $S_e$ and implicit symptom set $S_i$ as an example, where $|S_e| = n_e$ and $|S_i| = n_i$. We simulate $t$ dialog states based on $g_i$ with the following steps.
\begin{itemize}[noitemsep,topsep=0pt]
\item Select the entire explicit symptom set $S_e$.
\item Randomly select $n_i' \in [0, n_i)$ and sample $n_i'$ implicit symptoms to construct $S_i' \subset S_i$
\item Randomly select $n_u \in [0, T_{max} - n_i')$ and sample $n_u$ unrelated symptoms to construct set $S_u$. $T_{max}$ stands for the maximum number of symptoms can be queried.
\item Set the number of turns with $t = n_i' + n_u$.
\item If $n_i' = n_u = 0$, set AgentAction to ``Initiate''. Else set the AgentAction to ``Request''.
\item Randomly select a symptom $s \in S_i \cup S_u$. If $s \in S_u$, set UserAction to ``NotSure'', else set it to ``Confirm'' or ``Deny'' based on $g_i$.
\item Set current slot to $S_e \cup S_i' \cup S_u$.
\item Randomly select a implicit symptom $s_l \in S_i - S_i'$ as the prediction label.
\end{itemize}

\noindent \textbf{Action Prediction} We simulate dialog states for the dialog action prediction task with the same procedure as described above, except that we can involve all implicit symptoms. If all implicit symptoms are included, the training label will set to ``Conclude'', otherwise the label will be ``Query''.

We train MLPs and GMemNNs on both tasks after the training sets are generated. The models are trained with the simulated dialog states and labels with the stochastic gradient descent (SGD) algorithm.

\section{Experiments}
We train and evaluate our models on the Muzhi corpus. The symptom predictor and the dialog action predictor are trained separately. Using the same strategy of simulating the training set, we also generated test sets for symptom prediction and action prediction respectively using the test user goals with the same method. The generated test sets are used for evaluating the performances of our models on both unit tasks.

After evaluating the models in with the unit tasks, we conduct conversational evaluations using the trained models and a user simulator. We evaluate the performance of the models by accounting the number of implicit and unrelated symptoms queried in the conversations.

\subsection{Action Prediction}
For action prediction, we simulate 20 dialog states for each user goal in both training and test sets. All simulated states contain the entire explicit symptom sets. 10 of the 20 states also contain the complete implicit symptom sets, thus they are labeled with ``$1$'', meaning that the dialog system should conclude the dialog given these states in a dialog. The other states only contains a proper subset of implicit symptoms. These states are labeled with ``$0$'', meaning that the agent should continue querying symptoms. We have 11,360 training states and 2,840 test states.

We train an MLP and a GMemNN model on the simulated training sets. The MLP model has one hidden layer with 128 neurons, while the size of the hidden layers of GMemNN is set to 64. The models are trained with stochastic gradient descent (SGD) algorithm. The learning rate for training the MLP is 0.025, and is set to 0.035 for training the GMemNN. A weight decay rate of 0.001 is applied for training both models. Both models are trained for 40 epochs.

The experimental results are shown in Table \ref{tab:exp-unit}. All experimental results are obtained by running 5 independent experiments for each model from data simulation. The GMemNN model outperformed the MLP model with a small margin. The experimental results indicated that action prediction is not a hard classification task that external knowledge and complex neural networks do not help much.
\begin{table}[]
\centering
\begin{tabular}{@{}cccc@{}}
\toprule
\textbf{Unit Task}                                                                 & \textbf{Model} & \textbf{Acc} (\%) & \textbf{Stdv} (\%) \\ \midrule
\multirow{2}{*}{\begin{tabular}[c]{@{}c@{}}Action\\ Prediction\end{tabular}}  & MLP            & 94.14             & 0.27                \\
                                                                              & GMemNN         & 94.50             & 0.42                \\ \midrule
\multirow{2}{*}{\begin{tabular}[c]{@{}c@{}}Symptom\\ Prediction\end{tabular}} & MLP            & 45.10             & 0.62                \\
                                                                              & GMemNN         & 47.88             & 1.18                \\ \bottomrule
\end{tabular}
\caption{Unit task evaluation results of the action and symptom prediction tasks. Acc stands for average accuracy, and Stdv stands for the standard deviation of the accuracies. The statistics are obtained by running 10 experiments for each model on each task.}
\label{tab:exp-unit}
\end{table}

\subsection{Implicit Symptom Prediction}
For implicit symptom prediction, we simulate 10 dialog states for each user goal in both training and test sets. All dialog states contains the complete explicit symptom set and a proper subset of implicit symptoms. A random number of unrelated symptoms are also included. The label for training set is randomly sampled from implicit symptoms that are not included in the dialog state.

We train the neural networks for the implicit symptom prediction task with SGD. The architectures of MLP and GMemNN are the same as the models applied for action prediction respectively. We also apply the same hyper-parameter settings for training as the previous task.

The experimental results of symptom prediction are shown in Table \ref{tab:exp-unit}, which are also collected by runing 5 independent experiments from data simulation. The GMemNN model significantly outperformed the basic MLP model by $2.7\%$ on average and the performance is more stable. Comparing with the action prediction task, symptom prediction is much more difficult. As a result, domain specific knowledge can improve the performance more significantly.

\subsection{Conversational Evaluation}
We also evaluate our model by conducting dialogs using the original test split of user goals in the Muzhi corpus. For each test user goal, we generate a conversation using the dialog action predictor, the implicit symptom predictor, and a rule-based user simulator.

The user simulator initiates a dialog by providing a set of explicit symptoms as the initiate state. In each dialog step, the action predictor decides if the current state is informative enough to conclude the dialog. If a conclusion action is predicted, the system stops the conversation. Otherwise, the system queries the user simulator with a symptom selected by the symptom predictor. If the selected symptom is positive in the implicit symptom set, the user simulator confirms the query. If it is negative in the implicit symptom set, the user simulator denies the query. If the selected symptom is not in the implicit, the user simulator responses ``NotSure''. The dialog continues until the ``Conclusion'' action is selected, or the maximum limit of dialog turns is reached.

For each test user goal, we calculate the number of unrelated symptoms queried $N_u$, the number of dialog turns $N$, and the ratio of detected implicit $R_d$. Given the number of all implicit symptoms $N_i$ and the number of the detected implicit symptoms $N_i'$, we calculate the hit rate $R_h$, unrelated rate $R_u$, and the F1 score by
\begin{equation}
R_h = \frac{N_i'}{N_i}, \: R_u = \frac{N_u}{N}, \: F_1 = \frac{2 \* R_h \* (1 - R_u)}{R_h + 1 - R_u}
\end{equation}

We evaluate the models by calculating and comparing $R_d$, $R_u$, and F1 score averaged by the number of conversations. The experimental results are shown in Table \ref{tab:exp-diag}.
\begin{table}[]
\centering
\begin{tabular}{@{}cccc@{}}
\toprule
\textbf{Model} & \textbf{Hit} (\%)   & \textbf{UnRel} (\%) & \textbf{F1} (\%) \\ \midrule
MLP-AD         & 9.62          & 83.37 & 18.75                    \\
MLP-ASD        & 63.26          & 81.88 & 31.35                         \\
GMemNN         & \textbf{67.30} & \textbf{81.05} & \textbf{32.59}                         \\ \bottomrule
\end{tabular}
\caption{The experimental results of the conversational evaluation. MLP-AD stands for the pretrained state-of-the-art MLP model for automatic diagnosis (AD) provided by the authors of \citet{xu2019end}. MLP-ASD stands for the MLP model for automatic symptom detection (ASD) in this work. Hit stands for average hit rate $R_h$, UnRel stands for average unrelated rate $R_u$.}
\label{tab:exp-diag}
\end{table}

The experiments are conducted by setting the tolerate rate (TolR) to 10, meaning allowing the agent to query up to 10 symptoms. The experimental results showed that the MLP-ASD and GMemNN models detected significantly more implicit symptoms than the MLP-AD model \cite{xu2019end}, which makes diagnosis by querying only $9.62\%$ of implicit symptoms that a human doctor would ask about. Comparing the MLP-AD and GMemNN models, the GMemNN model significantly outperformed the MLP model by $4.04\%$ hit rate with $0.83\%$ lower unrelated rate. The improvement on F1 score is $1.24\%$.

We use tolerate rate (TolR) to limit the number of dialog turns. If the symptom predictor is completely random and the TolR equals to the number of symptoms, the hit rate $R_h$ will be $100\%$. However, querying all symptoms costs too much time for the patient. Since the average number of symptoms per user goal is $3.26$, the average unrelate rate $R_u$ of such a system will be $(66 - 3.26) / 66 = 95.06\%$ and the F1 score will be as low as $9.45\%$.
%
\begin{figure}[t]
\centering
\includegraphics[height=2.45in]{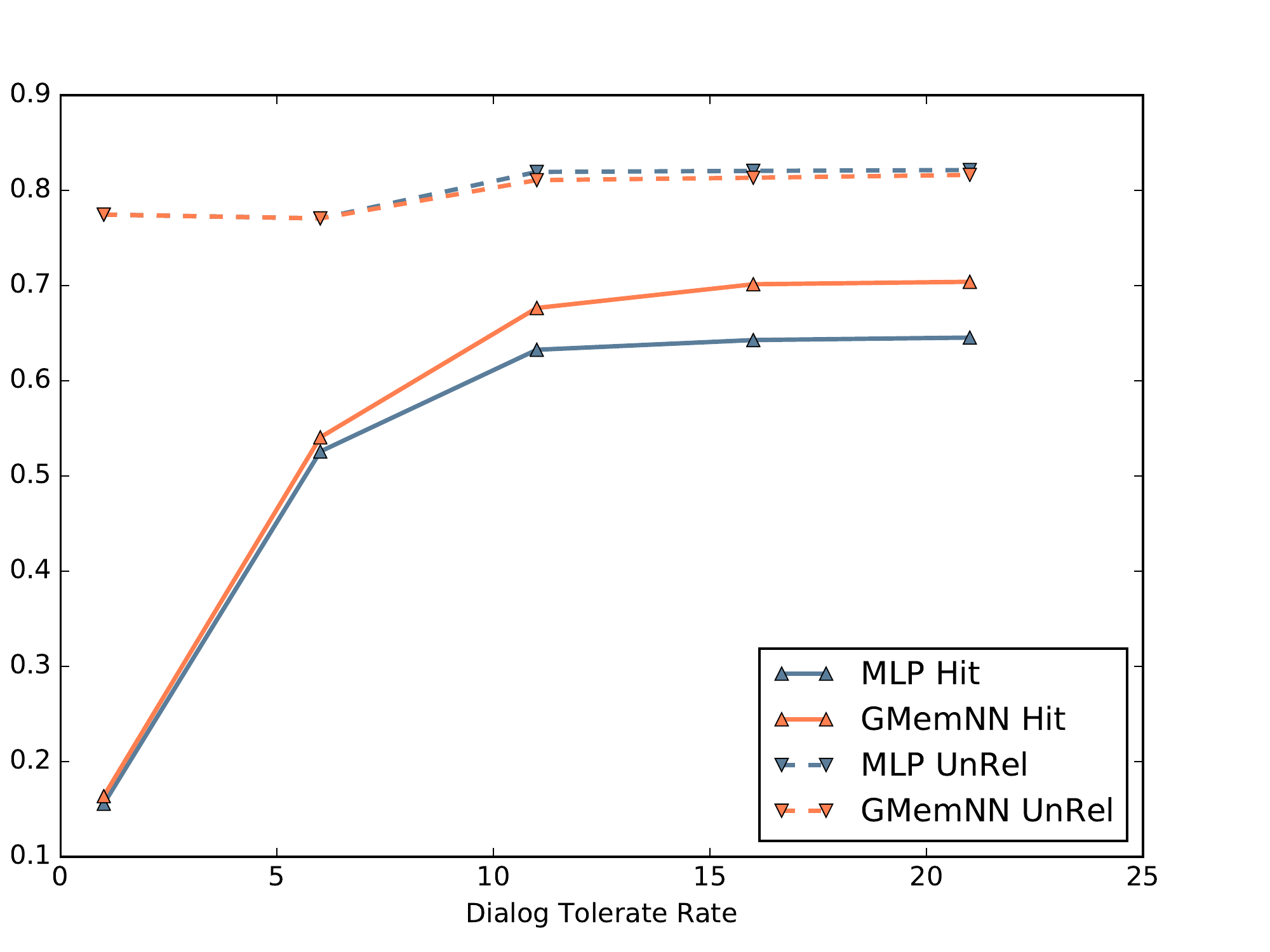}
\caption{The effect of tolerate rate on hit rate and unrelate rate for the MLP and the GMemNN models.}
\label{fig:tolr}
\end{figure}

To understand the effect of the tolerate rate, we visualized the relation between $R_h$, $R_u$, and TolR in Figure \ref{fig:tolr}. The plot indicates that increasing TolR from 1 to 10 can significantly improve the hit rates. However, the improvement vanishes after the 15th query because having too many queried symptoms makes the dialog states noisy. When the TolR is less than 10, the performance gap between The MLP and GMemNN model is not as large as the cases where TolR is larger than 10. There are two reasons for this phenomenon. I. some symptoms are queried by human doctors very frequently and they are equally easy for both models to predict; II. The GMemNN has better ability to model and process noisy inputs.

\section{Conclusion}
In this work, we propose a new task: detecting implicit symptoms of patient with an automatic dialog system. We construct the system with a dialog action prediction module and a symptom query module. We first implement and evaluate a baseline system based on multi-layer perceptrons (MLPs). To improve the performance of the system, we annotate a medical-domain knowledge graph and propose the graph memory network (GMemNN) model. We systematically evaluate and compare both models with unit tasks and conversations. We also studied how the number of dialog turns effects the performance of the systems. Experiments showed that both models can detect more than $60\%$ implicit symptoms using limited turns of dialogs, which significantly outperformed the state-of-the-art automatic diagnosis system. In future work, we will expand the knowledge graph and aim to assist human doctors by making the clinical interview process more efficient.

\bibliography{anthology,emnlp2020}
\bibliographystyle{acl_natbib}

\end{document}